\documentclass[10pt,twocolumn,letterpaper]{article}

\usepackage{iccv}
\usepackage{times}
\usepackage{epsfig}
\usepackage{graphicx}
\usepackage{amsmath}
\usepackage{amssymb}

\usepackage{booktabs}
\usepackage{subfigure}
\usepackage{color}
\usepackage{colortbl}
\usepackage{array} 
\usepackage{multirow}
\usepackage{enumitem}
\usepackage[accsupp]{axessibility}  % Improves PDF readability for those with disabilities.

\usepackage[pagebackref=true,breaklinks=true,letterpaper=true,colorlinks,bookmarks=false]{hyperref}

\iccvfinalcopy % *** Uncomment this line for the final submission

\newtheorem{Theo}{Theorem}

\newtheorem{Proof}{Proof}

\graphicspath{{fig/}}

\newcommand{\tablestyle}[2]{\setlength{\tabcolsep}{#1}\renewcommand{\arraystretch}{#2}\centering\footnotesize}
\renewcommand{\paragraph}[1]{\vspace{1.25mm}\noindent\textbf{#1}}
\newcommand{\app}{\raise.17ex\hbox{$\scriptstyle\sim$}}
\newcommand{\plus}{\raise.17ex\hbox{$\scriptstyle +$}}
\newcommand{\x}{{\times}}

\definecolor{baselinecolor}{gray}{.9}
\newcommand{\baseline}[1]{\cellcolor{baselinecolor}{#1}}

\begin{document}

%%%%%%%%% TITLE
\title{Graph Matching with Bi-level Noisy Correspondence}

\author{Yijie Lin\textsuperscript{\rm 1},
Mouxing Yang\textsuperscript{\rm 1}, Jun Yu\textsuperscript{\rm 2}, Peng Hu\textsuperscript{\rm 1}, Changqing Zhang\textsuperscript{\rm 3$*$}, Xi Peng\textsuperscript{\rm 1}\thanks{Corresponding author}
\\
\textsuperscript{\rm 1} Sichuan University,
\textsuperscript{\rm 2} Hangzhou Dianzi University, \textsuperscript{\rm 3} Tianjin University\\
{\tt\small \{linyijie.gm,yangmouxing,penghu.ml,pengx.gm\}@gmail.com, }
\\{\tt\small yujun@hdu.edu.cn, zhangchangqing@tju.edu.cn}
}

\maketitle
\ificcvfinal\thispagestyle{empty}\fi

%%%%%%%%% ABSTRACT

\begin{abstract}
In this paper, we study a novel and widely existing problem in graph matching (GM), namely, Bi-level Noisy Correspondence (BNC), which refers to node-level noisy correspondence (NNC) and edge-level noisy correspondence (ENC). In brief, on the one hand, due to the poor recognizability and viewpoint differences between images, it is inevitable to inaccurately annotate some keypoints with offset and confusion, leading to the mismatch between two associated nodes, \textit{i.e.}, NNC. On the other hand, the noisy node-to-node correspondence will further contaminate the edge-to-edge correspondence, thus leading to ENC. For the BNC challenge, we propose a novel method termed Contrastive Matching with Momentum Distillation. Specifically, the proposed method is with a robust quadratic contrastive loss which enjoys the following merits: i) better exploring the node-to-node and edge-to-edge correlations through a GM customized quadratic contrastive learning paradigm; ii) adaptively penalizing the noisy assignments based on the confidence estimated by the momentum teacher. Extensive experiments on three real-world datasets show the robustness of our model compared with 12 competitive baselines. The code is available at {\href{https://github.com/XLearning-SCU/2023-ICCV-COMMON}{https://github.com/XLearning-SCU/2023-ICCV-COMMON}}.
\end{abstract}

%%%%%%%%% BODY TEXT

\section{Introduction}
\label{sec:intro}

\begin{figure}[t]
  \centering
\includegraphics[width=1\columnwidth]{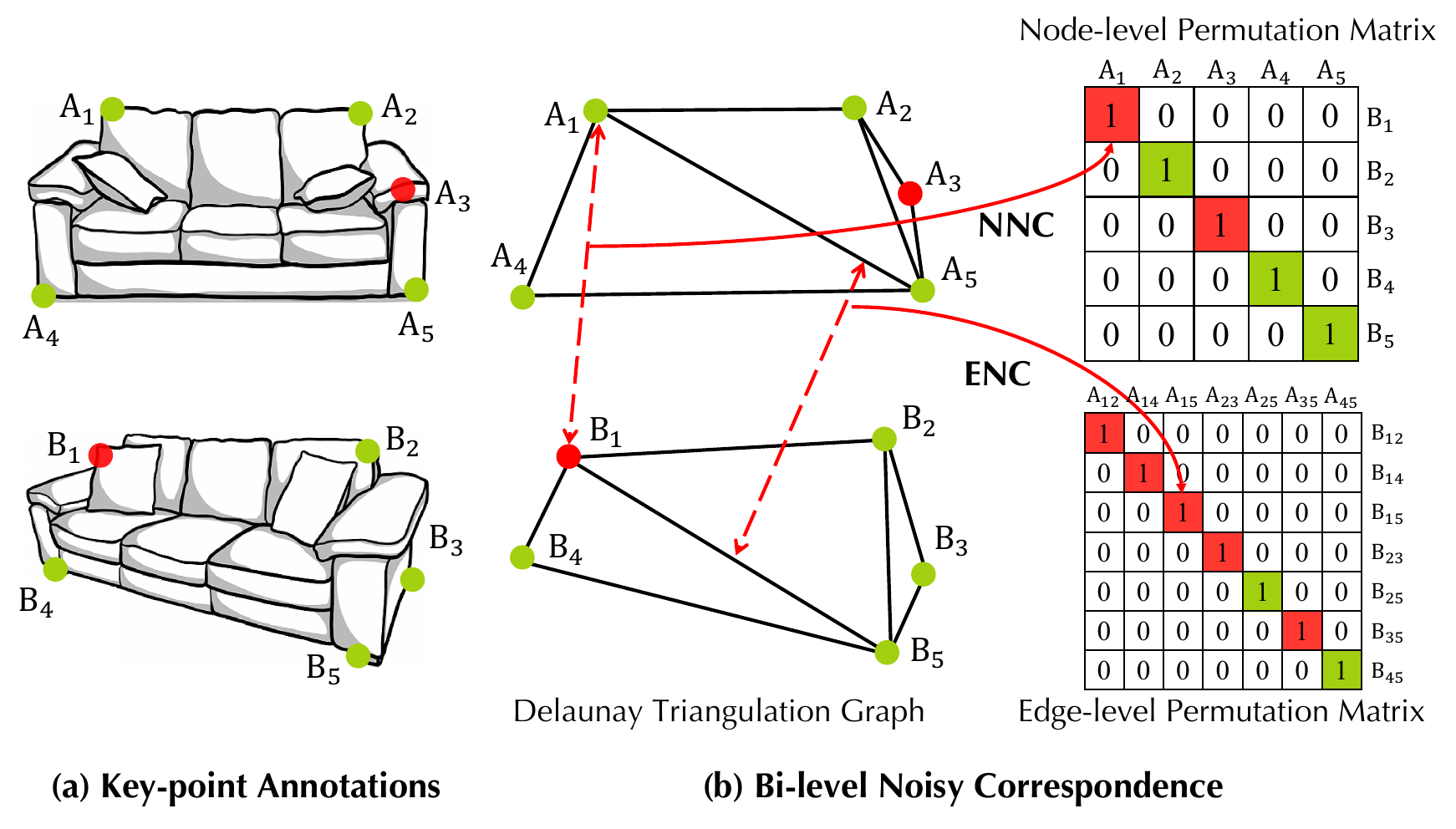}
 \caption{
 \textbf{An illustrative example of Bi-level Noisy Correspondence (BNC)}. The green and red dots denote the correctly and wrongly annotated keypoints. The green and red squares denote the true and false assignments.
 \textbf{(a)} Keypoint $\mathrm{A}_3$ and $\mathrm{B}_1$ are wrongly annotated due to the occlusion caused by the view-point difference, and the low recognizability, respectively. 
 \textbf{(b)}  Given the keypoints, GM methods construct the graph structure for further matching.
The matching procedure will inevitably encounter \textbf{BNC}, which refers to Node-level Noisy Correspondence (NNC) and Edge-level Noisy Correspondence (ENC). Specifically, \textbf{NNC} denotes the false matching between nodes, \textit{e.g.}, nodes $\mathrm{A}_1$ and $\mathrm{B}_1$ are wrongly matched. 
  \textbf{ENC} is accompanied with NNC as the edge weight is derived from the feature and position of nodes. Once the node is imperfectly annotated, the edges derived from it and the edges derived from its counterpart will be wrongly associated. 
  For instance, the edge between  $\mathrm{A}_{1}$ and  $\mathrm{A}_{5}$ (marked as $\mathrm{A}_{15}$) is wrongly associated with that between $\mathrm{B}_{1}$ and  $\mathrm{B}_{5}$ (marked as $\mathrm{B}_{15}$). 
   }
   \label{fig:basic}
\end{figure}

Graph Matching (GM)~\cite{cho2010reweighted,GMN} aims to establish correspondences between keypoints of different graphs, which plays an important role in various applications such as object tracking~\cite{yang2021learning,ufer2017deep}, scene graph discovery~\cite{chen2020graph}, Simultaneous Localization and Mapping~\cite{cadena2016past}, and Structure-from-Motion~\cite{superglue}. The key of GM is to explore and exploit the bi-level affinities between graphs, \textit{i.e.,} node-to-node (\textit{linear}) and edge-to-edge (\textit{quadratic}) affinity. For this purpose, existing methods have been devoted to integrating the bi-level information by designing advanced graph neural networks~\cite{PCA,IPCA,superglue,CIE} or differentiable quadratic loss~\cite{QCDGM,BBGM}. Based on the encoded high-order geometrical information, graph matching achieves promising results in correspondence estimation.

However, the success of existing GM methods highly depends on an implicit assumption, \textit{i.e.}, the annotated keypoint pairs are correctly associated. Unfortunately, in practice, the manual annotation process is susceptible to poor recognizability~\cite{pascalvoc} and viewpoint differences between images~\cite{spair71k}, which probably results in offset and even false keypoint annotations (Please refer to Fig.~\ref{fig:visualization} for examples). As shown in {Fig.~\ref{fig:basic}}, the inaccurate annotations will inevitably lead to \textbf{Bi-level Noisy Correspondence} (BNC) problem. Specifically, BNC refers to node-level noisy correspondence (NNC) and accompanied edge-level noisy correspondence (ENC). 
As shown in Fig.~\ref{fig:nc-a}, BNC leads to serious performance degradation due to two potential issues. 
On the one hand, BNC would hinder both node-level and edge-level representation learning. Specifically, representation learning on graphs requires information propagation and aggregation between the nodes and edges, therefore, BNC will cause accumulative errors and mislead the model optimization.
On the other hand, GM is subject to one-to-one mapping constraints, \textit{i.e.}, each keypoint in the first graph must have a unique correspondence in the second graph.  Hence, ``\textit{a slight move in one part may affect the situation as a whole}", \textit{i.e.}, one single noisy correspondence pair might result in global alignment failure and even affect the optimization of correctly annotated pairs.

To the best of our knowledge, such an essential problem has not been touched so far and it is quite challenging to solve it due to the following two reasons:  
i) without prior engineering like verification labels~\cite{lee2018cleannet} for noisy correspondence, it is nearly impossible to pre-process them before training, \textit{i.e.}, correctly distinguish and discard the noisy correspondence in advance. Hence, it is highly expected to develop a robust matching method. ii) however, GM is not a simple one-to-many optimization problem (\textit{e.g.}, classification) but a many-to-many combinatorial optimization problem. One cannot solve the BNC challenge by trivially resorting to label noise learning methods~\cite{Iscen_2022_CVPR,Dividemix,yang2022DART} as they only enjoy robustness on one-to-many classification tasks.

To explore an effective solution to this challenge, we propose a novel method termed COntrastive Matching with MOmentum distillatioN (COMMON). Specifically, COMMON is equipped with a robust quadratic contrastive loss that incorporates both linear and quadratic geometrical information into the contrastive learning paradigm~\cite{simclr,moco,wang2020understanding}.
More specifically, this study is motivated by ~\cite{moskalev2022contrasting} that the vanilla contrastive loss is equivalent to the linear assignment from the combinatorial optimization theory. In other words, the quadratic structure information could be incorporated into contrastive learning. Based on this motivation, we endow the contrastive loss with quadratic information through two novel graph-geometric consistency regularizers. 
To enhance the robustness against BNC, the proposed loss adaptively penalizes the noisy correspondence through a   momentum distillation strategy based on the \textit{memorization effect} of deep neural networks~\cite{arpit2017closer,xia2020robust,huang2021learning}, \textit{i.e.}, deep networks are apt to learn the simple patterns before fitting the noise. Motivated by this empirical observation, we keep a momentum version~\cite{albef,moco} of the GM model by taking the moving average of its parameters during training. The momentum teacher model could generate high-quality pseudo-targets as additional supervision on the quadratic contrastive loss without resorting to extra verification labels. With bi-level distillation on both node and edge alignment, the proposed robust quadratic contrastive loss effectively mitigates the negative affect of BNC.

The main contributions of this work are:
\begin{itemize}
[topsep=3pt,itemsep=-1pt]
\item We reveal a new challenge for graph matching, termed bi-level noisy correspondence (BNC). BNC refers to  noisy correspondence on both node and edge levels, which would lead to serious performance degradation. 
\item To tackle the BNC challenge, we propose a robust quadratic contrastive learning loss that utilizes the momentum distillation strategy to rebalance the noisy assignment. The proposed contrastive loss explores the quadratic information through two simple graph consistency regularizers.
\item Extensive experiments on real-world data verify the effectiveness of our method against noisy correspondence. On Pascal VOC, Spair-71k and Willow, we achieve absolute improvements of 1.6\%, 1.4\%, and 1.4\% compared to the state-of-the-art methods.
\end{itemize}

\section{Related work}
In this section, we briefly introduce some recent developments in deep graph matching and contrastive learning.

\subsection{Deep Graph Matching}
Deep GM~\cite{GMN,DGMC} aims at aligning the associated keypoints from different graphs according to node-to-node and edge-to-edge correlations. 
To achieve better matching effects, existing methods mainly focus on utilizing the high-order information in the graph structures. According to the choice of exploiting high-order information, most existing methods could be divided into the following two groups. i) network-designed based methods~\cite{LCS,CIE,liu2021joint,jiang2021gamnet} which implicitly aggregate the high-order information through GM-customized networks. For instance, PCA~\cite{PCA} employs graph convolutional networks to aggregate inner-graph and cross-graph structure information, and NGM~\cite{NGM} proposes a matching-aware graph convolution scheme with Sinkhorn iteration~\cite{sinkhorn}. ii) loss-designed based methods~\cite{liu2021stochastic,QCDGM,BBGM} that explicitly learn the high-order information through different differentiable quadratic loss or optimization strategies. For instance, QCDGM~\cite{QCDGM} modifies the Frank-Wolfe algorithm into a differentiable optimization scheme for quadratic constraint, and BBGM~\cite{BBGM} proposes a differentiable combinatorial solver for quadratic assignment. 

Although deep GM has achieved promising performance, almost all existing methods assume that the node-to-node and edge-to-edge correspondence is faultless and correctly associated. 
However, in practice, the assumption is daunting and nearly impossible to satisfy as aforementioned in Introduction, \textit{i.e.}, BNC is inevitable due to poor annotations. Note that, although some efforts have been devoted to achieving robust GM, these methods mainly focus on the robustness against outliers~\cite{wang2020zero,superglue,qu2021adaptive,BBGM} and adversarial attack~\cite{ASAR}, which is remarkably different from the revealed BNC challenge. To achieve robustness against BNC, this study employs a momentum distillation strategy that rebalances the matching loss in a data-driven way. To the best of our knowledge, this study could be the first work on BNC-robust graph matching.

\subsection{Contrastive Learning}
As one of the most effective representation learning paradigms, contrastive learning~\cite{9852291,wang2021dense,lin2022scc,xie2021propagate,li2022twin,goel2022cyclip,SimCSE} attempts to align representations of the data from different views~\cite{completer,YangXZCH19,JiangXYCH19,o2020unsupervised} by maximizing the similarity between positive pairs and minimizing that of negative pairs.
The major difference between these works lies in the strategy of constructing positive and negative pairs.
For example, SimCLR~\cite{simclr} constructs positive samples through data augmentation and takes different mini-batch images as negatives. Moco~\cite{moco} proposes a memory bank module that stores a large number of negatives to increase the contrasting effect.

The differences between this study and existing works are two-fold. On the one hand, most existing methods assume that positive pairs are closely associated, which is hard to satisfy in practical applications. 
In contrast, our momentum distillation solution improves the robustness of contrastive learning against imperfect correspondence. 
On the other hand, as pointed by~\cite{moskalev2022contrasting}, almost all contrastive methods only consider instance discrimination problem, \textit{i.e.}, measuring the alignment between individual pairs of objects while ignoring the high-order correlation among these objects. To tackle this problem, we propose a novel quadratic contrastive learning loss that not only aligns the objects (node) but also aligns the correlation between objects (edge). Different from the quadratic loss proposed in~\cite{moskalev2022contrasting}, our quadratic term is implemented with the stable mean square error instead of singular value decomposition. In addition, our quadratic term incorporates the extra cross-graph quadratic information, whereas \cite{moskalev2022contrasting} does not. What is more distinct, our loss is designed to achieve the robustness against the BNC problem. Experiments demonstrate that the proposed quadratic loss improves the performance of contrastive learning.

\begin{figure*}
  \centering
  \includegraphics[width=2\columnwidth]{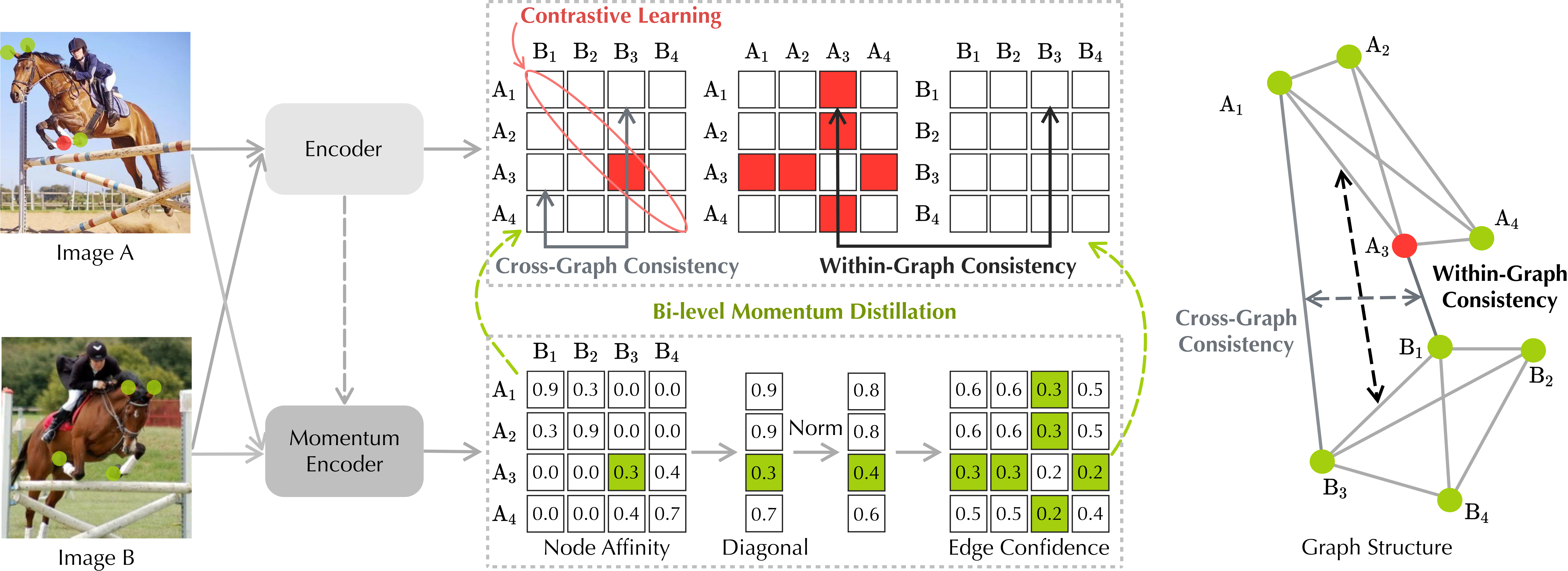}
  \caption{\textbf{Overview of COMMON.} In the figure, the green and red dots denote the correctly and wrongly annotated keypoints. 
  The red and green squares indicate the noisy correspondence and corresponding estimated confidence.
  By feeding a pair of images into the encoder, we obtain the node similarity matrix and two edge adjacency matrices. To better explore the bi-level correlations, we propose a quadratic contrastive loss which consists of three jointly learning objectives, \textit{i.e.}, contrastive learning loss, cross-graph consistency loss, and within-graph consistency loss. To mitigate the influence of BNC, we further feed the input into the momentum encoder that adaptively rebalances the noisy correspondence based on the estimated node and edge confidences.
  }
  \label{fig:framework}
\end{figure*}

\section{Method}
In this section, we first introduce the problem definition (Section~\ref{sec:3.1}). Then we delineate how the proposed quadratic contrastive learning loss better explores the node-to-node and edge-to-edge correlations (Section~\ref{sec:3.2}), and how the designed loss adaptively penalizes the noisy correspondence through momentum distillation (Section~\ref{sec:3.3}).

\subsection{Problem Definition}
\label{sec:3.1}
For two images with $n$ and $m$ keypoints each ($n\leq m$), graph matching aims to build the  node-to-node correspondence between two given graphs $\mathcal{G}_A$ and $\mathcal{G}_B$ built based on these keypoints. Let $\mathbf{V}_A\in\mathbb{R}^{n\times d}$ and $\mathbf{V}_B\in\mathbb{R}^{m\times d}$ denote feature matrix of keypoints in $\mathcal{G}_A$ and $\mathcal{G}_B$, respectively, where each row of $\mathbf{V}_A$ and $\mathbf{V}_B$ is a feature vector of a keypoint. $\mathbf{F}_A=\mathbf{V}_A\mathbf{V}_A^{\top} \in \mathbb{R}^{n \times n}, \mathbf{F}_B =\mathbf{V}_B\mathbf{V}_B^{\top}\in \mathbb{R}^{m \times m}$ indicate the adjacency matrices which encode the edge information in graphs $\mathcal{G}_A$ and $\mathcal{G}_B$.
Formally, graph matching seeks to solve,
\begin{equation}
\min L_{\mathrm{Y}} \left(\mathbf{Y}_{gt},\mathbf{Y}\right),
\label{eq:ly}
\end{equation}
where $L_{\mathrm{Y}}$ measures the discrepancy between the ground-truth assignment $\mathbf{Y}_{gt}$ and the matching result $\mathbf{Y}$, \textit{e.g.}, cross-entropy loss~\cite{NGM,ASAR} and hamming distance loss~\cite{BBGM,selfgm}. The matching result $\mathbf{Y}$ is generally obtained by,
\begin{equation}
\begin{aligned}
\underset{\mathbf{Y}}{\operatorname{argmax} } & \quad
 \underbrace{ \operatorname{tr}\left(\mathbf{Y}^{\top} \mathbf{F}_A \mathbf{Y} \mathbf{F}_B\right)}_{\text {Quadratic edge affinity}}
 +\underbrace{\operatorname{tr}\left(\mathbf{S}^{\top} \mathbf{Y}\right)}_{\text{Linear node affinity}}  \\
  s.t. &\quad   \mathbf{Y} \in\{0,1\}^{n \times m}, ~ \mathbf{Y} \mathbf{1}_m=\mathbf{1}_n, ~
   \mathbf{Y}^{\top} \mathbf{1}_n \leq \mathbf{1}_m
  \end{aligned}
  \label{eq:1}
\end{equation}
where $\mathbf{S}=\mathbf{V}_A \mathbf{V}_B^{\top} \in \mathbb{R}^{n \times m}$ denotes node-to-node similarity matrix. 
By combining Eqs.~(\ref{eq:ly}) and (\ref{eq:1}), it shows that GM encourages to learn better node and edge features that leads to correct assignment.

However, due to poor recognizability and viewpoint difference of images, it is extremely hard and time-consuming to precisely annotate keypoints, leading to \textbf{incorrect} assignment $\mathbf{Y}_{gt}$. 
As shown in Fig.~\ref{fig:basic} and Eq.~(\ref{eq:1}), the incorrect $\mathbf{Y}_{gt}$ induces false matching on both node and edge levels, \textit{i.e.}, bi-level noisy correspondence.  
Accordingly, the BNC problem tends to cause false optimization on both node and edge levels, leading to serious performance degradation. As aforementioned, it is hard to preprocess the noisy correspondence before training. Therefore, our method attempts to explore a robust training strategy which could effectively mitigate the negative impact of BNC.

\subsection{Quadratic Contrastive Learning for Graph Matching}
\label{sec:3.2}

In this section, we propose the quadratic contrastive learning loss for graph matching. Specifically, based on the combinatorial optimization theory~\cite{moskalev2022contrasting} that contrastive learning is equivalent to linear assignment problem, we accordingly introduce two graph consistency regularizers that endow contrastive learning with quadratic information.

As pointed by~\cite{moskalev2022contrasting}, the linear assignment problem (the second term in Eq.~(\ref{eq:1})) could be formulated by structured linear loss~\cite{tsochantaridis2005large}:
\begin{equation}
L= \max_{\mathbf{Y} \in \Pi} \operatorname{tr}\left(\mathbf{S} \mathbf{Y}^{\top}\right)
-\operatorname{tr}\left(\mathbf{S} \mathbf{Y}_{g t}^{\top}\right),
 \label{eq:linear}
\end{equation}
where $\Pi$ denotes a set containing all $n\times m$ permutation matrices that follows the constraint in Eq.~(\ref{eq:1}).
Note that $L\ge0$ and $L= 0$ i.f.f. the node similarities produced by the encoder lead to the correct assignment. Through minimizing the objective in Eq.~(\ref{eq:linear}), the encoder expected to correctly assign keypoints from one image to another.

To make the structure linear loss calculable, one could relax the constraint $\Pi$ with binary row-stochastic relaxation ($[\mathbf{Y} ]_{ij}\in\{0,1\}$ and $\sum_j\mathbf{Y}_{i j}=1 ~ \forall i$) and utilize the common log-sum-exp approximation~\cite{beck2012smoothing} on the max function of Eq.~(\ref{eq:linear}). Then the linear assignment loss could be reformulated as the InfoNCE contrastive loss as proved by~\cite{moskalev2022contrasting}:
\begin{equation}
  \begin{aligned}
L =- \sum_{(i, j) \in \mathbf{Y}_{g t}}[\mathbf{S}]_{i j}+\tau \sum_i \log (\sum_j \exp (\frac{1}{\tau}[\mathbf{S}]_{i j})),
\end{aligned}
\label{eq:loglinear}
\end{equation}
where $\tau$ controls the degree of smoothness. 
 For our case, we reformulate the equivalence theory between the linear assignment and contrastive learning as follows. The detailed proof please refer to Appendix~\ref{app:a}.
\begin{Theo}
The log-sum-exp ~\cite{beck2012smoothing} smoothed structured linear assignment loss $L$ with row-stochastic relaxation is equivalent to the InfoNCE contrastive loss~\cite{moco,mocov2}.
\label{the:1}
\end{Theo}

Notably, the above theorem indicates that contrastive learning Eq.~(\ref{eq:loglinear}) is a fast and differentiable approximation to the linear assignment problem thus could be efficiently used to align the keypoints in deep graph matching directly.
To simply conduct contrastive learning for graph matching, we align the keypoints $\mathbf{V}_A$ and $\mathbf{V}_B$ according to $\mathbf{Y}_{gt}$ and only remain the keypoints which have the counterpart for training, obtaining aligned keypoints $\mathbf{P}_A,\mathbf{P}_B\in \mathbb{R}^{n\times d}$ for graph $\mathcal{G}_A$ and $\mathcal{G}_B$, respectively. Then we conduct contrastive learning on both rows and columns of the node similarity matrix $\mathbf{S}$ as~\cite{radford2021learning}:
\begin{equation}
\begin{aligned}
{L}_{\mathrm {InfoNCE }}=\mathcal{H}\left(\mathbf{I}_{n}, \rho\left(\mathbf{P}_A \mathbf{P}_B^{\top}\right)\right)+\mathcal{H}\left(\mathbf{I}_{n}, \rho\left(\mathbf{P}_B \mathbf{P}_A^{\top}\right)\right),
\end{aligned}
\label{eq:infonce}
\end{equation}
where $\mathbf{I}_{n}$ is the identity matrix, $\mathcal{H}$ is the row-wise cross-entropy function with mean reduction and $\rho$ is the softmax function applied row-wise such that each row sums to one, \textit{i.e.},
\begin{equation}
[\rho\left(\mathbf{P}_A \mathbf{P}_B^{\top}\right)]_{ij} =   
\frac{\exp ([\mathbf{P}_{A}]_i  [\mathbf{P}_{B}]_j^\top / \tau)}{\sum_{k=1}^n \exp \left([\mathbf{P}_{A}]_i [\mathbf{P}_{B}]^\top_k / \tau\right)},
\end{equation}
where $[\mathbf{P}_{A}]_i\in \mathbb{R}^{1\times d}$ is the $i$-row of $\mathbf{P}_{A}$, and $\tau$ is the softmax temperature fixed as $0.07$.

Although the popular InfoNCE loss might be an effective solution to the linear assignment problem, it ignores another essential perspective in graph matching, \textit{i.e.}, edge alignment. 
In fact, it is generally accepted that considering the edge information in graphs makes the matching more robust~\cite{PCA}.
Therefore, contrastive learning can not sufficiently utilize the graph structure and might obtain suboptimal results. Hence, we augment the contrastive loss with two novel graph-geometric consistency regularizers, namely, within-graph consistency and cross-graph consistency, to further exploit the edge correlation.

\paragraph{Within-graph consistency} aims to encourage the alignment between edges within each image,
\begin{equation}
  {L}_{\mathrm{within}}= \left \| \mathbf{P}_{A} \mathbf{P}_{A}^\top- \mathbf{P}_{B} \mathbf{P}_{B}^\top \right\|_F^2,
    \label{eq:within}
\end{equation}
where $\|\cdot\|_F$ is the Frobenius norm. The within-graph consistency is a direct quadratic assignment regularizer.

\paragraph{Cross-graph consistency} encourages the alignment of the edges across two images: 
\begin{equation}
  {L}_{\mathrm{cross}}=
   \left \| \mathbf{P}_{A} \mathbf{P}_{B}^\top- \mathbf{P}_{B} \mathbf{P}_{A}^\top \right\|_F^2.
  \label{eq:cross}
\end{equation}

For clarity, we provide a typical example in Fig.~\ref{fig:framework} to illustrate the above formulation. 
Intuitively, Eq.~(\ref{eq:within}) minimizes the difference between the edge in image A and its counterpart in image B, \textit{e.g.}, $\mathrm{A}_1\mathrm{A}_3$ and $\mathrm{B}_1\mathrm{B}_3$.
Apart from the alignment between within-graph edges, it is also encouraged to align the cross-graph edges.  
Specifically, to better align the keypoints between graphs, the semantic difference between objects needs to be eliminated, \textit{e.g.}, the difference between two kinds of horses. Hence, given two associated within-graph edges (\textit{e.g.}, $\mathrm{A}_1\mathrm{A}_3$ and $\mathrm{B}_1\mathrm{B}_3$), it is highly expected that the corresponding keypoints are equivalent and also exchangeable (\textit{e.g.}, $\mathrm{A}_3$ and $\mathrm{B}_3$). 

Therefore, we establish the cross-graph edges from within-graph edges by exchanging the counterpart of each keypoint and then minimize the difference between them (\textit{e.g.}, $\mathrm{A}_1\mathrm{B}_3$ and $\mathrm{B}_1\mathrm{A}_3$). Mathematically, Eq.~(\ref{eq:cross}) makes the matrix $\mathbf{S}$ symmetric to encourage the alignment between cross-graph edges. 
Although~\cite{PCA,superglue} propose the cross-graph neural network to encode the information from both graphs, they only propagate the information from one node in image $A$ ($B$) to other nodes in image $B$ ($A$). In contrast, our cross-graph consistency offers a more explicit quadratic supervision to ensure the cross-graph semantic consistency.

Notably, the above two geometric consistency terms are elegantly implemented as simple regularizers on contrastive learning and facilitate the matching procedure. Accordingly, the quadratic contrastive loss for graph matching is given as:
\begin{equation}
  L_{\mathrm{quadratic}}= L_{\mathrm{InfoNCE}} +{L}_{\mathrm{within}} +{L}_{\mathrm{cross}}.
\end{equation}

\begin{table*}
\tablestyle{1.6pt}{1.01}
{
\begin{tabular}{l|cccccccccccccccccccc|c}
\toprule
Method                       & {  Aero}                       & {  Bike}                       & {  Bird}  & {  Boat}  & {  Bottle} & {  Bus}   & {  Car}   & {  Cat}   & {  Chair} & {  Cow}   & {  Table} & {  Dog}   & {  Horse} & {  Mbike} & {  Person} & {  Plant} & {  Sheep} & {  Sofa}  & {  Train} & {  Tv}    & {  Mean}  \\
\midrule

GMN~\cite{GMN}     & 41.6 & 59.6 & 60.3 & 48.0 & 79.2 & 70.2 & 67.4 & 64.9 & 39.2 & 61.3 & 66.9 & 59.8 & 61.1 & 59.8 & 37.2 & 78.2 & 68.0 & 49.9 & 84.2 & 91.4 & 62.4 \\
PCA~\cite{PCA}  & 49.8 & 61.9 & 65.3 & 57.2 & 78.8 & 75.6 & 64.7 & 69.7 & 41.6 & 63.4 & 50.7 & 67.1 & 66.7 & 61.6 & 44.5 & 81.2 & 67.8 & 59.2 & 78.5 & 90.4 & 64.8 \\
NGM~\cite{NGM}     & 50.1 & 63.5 & 57.9 & 53.4 & 79.8 & 77.1 & 73.6 & 68.2 & 41.1 & 66.4 & 40.8 & 60.3 & 61.9 & 63.5 & 45.6 & 77.1 & 69.3 & 65.5 & 79.2 & 88.2 & 64.1 \\
IPCA~\cite{IPCA} & 53.8 & 66.2 & 67.1 & 61.2 & 80.4 & 75.3 & 72.6 & 72.5 & 44.6 & 65.2 & 54.3 & 67.2 & 67.9 & 64.2 & 47.9 & 84.4 & 70.8 & 64.0 & 83.8 & 90.8 & 67.7 \\

LCS~\cite{LCS} &46.9 &58.0& 63.6& 69.9 &87.8 &79.8& 71.8 &60.3 &44.8 &64.3& 79.4& 57.5& 64.4 &57.6 &52.4 &96.1 &62.9& 65.8& 94.4& 92.0 &68.5\\
CIE~\cite{CIE}   & 52.5 & 68.6 & 70.2 & 57.1 & 82.1 & 77.0 & 70.7 & 73.1 & 43.8 & 69.9 & 62.4 & 70.2 & 70.3 & 66.4 & 47.6 & 85.3 & 71.7 & 64.0 & 83.9 & 91.7 & 68.9 \\
QC-DGM~\cite{QCDGM} &49.6& 64.6& 67.1 &62.4& 82.1& 79.9 &74.8& 73.5& 43.0 &68.4 &66.5 &67.2 &71.4 &70.1 &48.6 &92.4& 69.2& 70.9 &90.9 &92.0 &70.3\\
DGMC~\cite{DGMC} & 50.4& 67.6& 70.7& 70.5& 87.2 &85.2& 82.5 &74.3 &46.2 &69.4 &69.9 &73.9 &73.8 &65.4 &51.6 &98.0 &73.2 &69.6 &94.3 &89.6 & 73.2
\\
BBGM~\cite{BBGM}    & 61.9 & 71.1 & \underline{79.7} & 79.0 & 87.4 & \underline{94.0} & \underline{89.5} & \underline{80.2} & 56.8 & 79.1 & 64.6 & 78.9 & 76.2 & 75.1 & \underline{65.2} & \underline{98.2} & 77.3 & {77.0} & 94.9 & \textbf{93.9} & 79.0 \\

NGM-v2~\cite{NGM}  & 61.8 & 71.2 & 77.6 & 78.8 & 87.3 & 93.6 & 87.7 & 79.8 & 55.4 & 77.8 & \underline{89.5} & 78.8 & \underline{80.1} & \underline{79.2} & 62.6 & 97.7 & 77.7 & 75.7 & 96.7 & 93.2 & 80.1 \\

SCGM~\cite{selfgm}&
\underline{62.9} &	72.9 &	79.6 &	\underline{79.5} &	\textbf{89.3} 	&\textbf{94.1} 	&89.1 	&79.2 &	58.4 	&79.3 &	80.5 &	\underline{79.9} 	&79.5 &	76.8 	&64.8 	&98.1 &	\underline{78.0} &	75.9 &	98.0 &	93.2 	&80.5
 \\

ASAR~\cite{ASAR} 
&\underline{62.9} &	\underline{74.3} &	79.5 	&\textbf{80.1} &	89.2 	&\underline{94.0} 	&88.9 	&78.9 &	\underline{58.8} 	&\underline{79.8} 	&88.2 &	78.9 &	79.5 	&77.9 &	64.9 	&\underline{98.2} 	&77.5 &	\underline{77.1} &	\underline{98.6} 	&93.7 &\underline{81.1} \\
\midrule

COMMON&
\textbf{65.6}&	\textbf{75.2}	&\textbf{80.8}&	\underline{79.5}	&\textbf{89.3}	&{92.3}	&\textbf{90.1}	&\textbf{81.8}	&\textbf{61.6}	&\textbf{80.7}	&\textbf{95.0}	&\textbf{82.0}	&\textbf{81.6}&	\textbf{79.5}&	\textbf{66.6}&	\textbf{98.9}	&\textbf{78.9}&	\textbf{80.9}	&\textbf{99.3}&	\underline{93.8}&	\textbf{82.7} \\

\bottomrule
\end{tabular}
}
\caption{
Keypoint matching accuracy (\%) on Pascal VOC  with standard intersection filtering. The best and second-best results are \textbf{highlighted} and \underline{underlined}, respectively.   }
\label{tab:pascal}
\end{table*}

\subsection{Momentum Distillation for Robust Matching}
\label{sec:3.3}

As aforementioned, bi-level noisy correspondence would cause failure on both node-to-node and edge-to-edge matching, resulting in serious performance degradation. Different from the traditional noisy label problem, BNC is a many-to-many combinatorial optimization problem rather than a one-to-many classification problem. It is inapplicable to distinguish~\cite{lee2018cleannet,karim2022unicon} or rectify~\cite{Dividemix} the noisy correspondence through existing noisy label classification methods.

Fortunately, Bengio \textit{et al.}~\cite{arpit2017closer} have empirically found the memorization effect of the deep neural networks, \textit{i.e.}, networks are apt to fit the simple patterns first. To be specific, precisely annotated correspondence could be regarded as simple patterns while noisy correspondence could be treated as complex ones.  
Motivated by this effect, we propose to learn from the high-quality pseudo-target generated by the momentum encoder~\cite{moco,albef} as shown in Fig~\ref{fig:framework}. The momentum encoder is a continuously-evolving teacher, which will keep the memory at the simple patterns to some extent by taking the exponential-moving-average of the parameters from the base encoder. Formally, denoting the parameters of the base encoder as $\theta_\mathrm{q}$ and those of momentum encoder as $\theta_\mathrm{k}$, we update $\theta_\mathrm{k}$ by:\begin{equation}
  \theta_{\mathrm{k}} \leftarrow t \theta_{\mathrm{k}}+(1-t) \theta_{\mathrm{q}},
\end{equation}
where $t$ is the momentum coefficient fixed as $0.995$ in our experiments. 
During training, we utilize the matching prediction of the momentum encoder to rebalance the proposed quadratic contrastive learning loss. Specifically, we propose a bi-level distillation on both node and edge levels with respect to contrastive loss and graph consistency loss.

\paragraph{Distillation on node level.}
We first compute the node features ${\mathbf{\hat{P}}}_{A}$ and ${\mathbf{\hat{P}}}_{B}$ from the momentum encoder. 
Then we introduce a new cross-entropy term into Eq.~(\ref{eq:infonce}) to match the alignment results between base encoder and the soft target of its momentum teacher~\cite{albef}. Formally,
\begin{equation}
\resizebox{1\linewidth}{!}{$
\begin{aligned}
&  L_{\mathrm{RInfoNCE}} = (1-\alpha)
  \left(\mathcal{H}\left(\mathbf{I}_{n}, \rho\left(\mathbf{P}_A \mathbf{P}_B^{\top}\right)\right)+\mathcal{H}\left(\mathbf{I}_{n}, \rho\left(\mathbf{P}_B \mathbf{P}_A^{\top}\right)\right)\right) 
\\ &+ {\alpha}\left(\mathcal{H}\left(
\rho\left(\mathbf{\hat{P}}_A \mathbf{\hat{P}}_B^{\top}\right),
{\rho}
\left(\mathbf{P}_A \mathbf{P}_B^{\top}\right) \right)
   +\mathcal{H}\left(
   \rho \left(
   \mathbf{\hat{P}}_B \mathbf{\hat{P}}_A^{\top}\right),
  \rho \left(\mathbf{{P}}_B \mathbf{{P}}_A^{\top}\right)
   \right)\right),
  \end{aligned}
     $}
     \label{eq:rinfonce}
\end{equation}
where $\alpha$ is the distillation temperature. 
The second term in Eq.~(\ref{eq:rinfonce}) allows the teacher to re-calibrate the keypoint alignment by replacing the impractical perfect alignment target $\mathbf{I}_n$ in Eq.~(\ref{eq:infonce}) with estimated soft-alignment probability.
To keep the stability in the preliminary training stage, $\alpha$ is linearly increasing from 0 to 0.4 in the first epoch and fixed to 0.4 afterward. As shown in Table~\ref{tab:parameter}, our method is insensitive to the choice of $\alpha$.

\paragraph{Distillation on edge level.}
As the edge-level noisy correspondence is accompanied with node-level noisy correspondence, we first estimate the confidence of each node and then estimate that of the edge based on the confidence of its two vertices. Formally, let $\textbf{s} \in \mathbb{R}^n$ denote the confidence of nodes where
\begin{equation}
\resizebox{1\linewidth}{!}{$
  \textbf{s}_i = \frac{1}{2}
  \left(\left[\mathbf{\hat{P}}_A \mathbf{\hat{P}}_B^{\top}\right]_{ii}
  \bigg/
   \sum_{j}^n \left[\mathbf{\hat{P}}_A \mathbf{\hat{P}}_B^{\top}\right]_{ij}+\left[\mathbf{\hat{P}}_A \mathbf{\hat{P}}_B^{\top}\right]_{ii}
   \bigg/ \sum_{j}^n
    \left[\mathbf{\hat{P}}_A \mathbf{\hat{P}}_B^{\top}\right]_{ji}\right).
$}
\end{equation}
The confidence of each edge is estimated by accumulating the confidence of its two vertexes, \textit{i.e.}, $\mathbf{W}=\sqrt{\textbf{s}  \otimes \textbf{s}}$ where $\otimes$ is the outer product. Based on the edge confidence $\mathbf{W}$, we re-weight the graph consistency loss (Eqs.~(\ref{eq:within}) and (\ref{eq:cross})) as follows,
\begin{equation}
\resizebox{1\linewidth}{!}{$
{L}_{\mathrm{Rgraph}}=
\left \|\mathbf{W}\circ ( \mathbf{P}_{A} \mathbf{P}_{A}^\top- \mathbf{P}_{B} \mathbf{P}_{B}^\top )\right\|_F^2
+  
   \left \| \mathbf{W} \circ( \mathbf{P}_{A} \mathbf{P}_{B}^\top- \mathbf{P}_{B} \mathbf{P}_{A}^\top) \right\|_F^2,
$}
\label{eq:rgraph}
\end{equation}
where $\circ$ is the Hadamard product.

Different from most existing distillation methods that acquire knowledge from a pre-trained model~\cite{hinton2015distilling,touvron2021training}, our momentum distillation strategy owns the following two merits: i) enjoying the robustness against BNC by exploiting the memory effect characteristic of the momentum model, \textit{i.e.}, exponential-moving-average parameters would resist overfitting the noise; ii) bootstrapping the matching performance without resorting to extra knowledge or models.

Finally, by combining Eqs.~(\ref{eq:rinfonce}) and (\ref{eq:rgraph}), the overall robust quadratic contrastive learning loss is induced as,
\begin{equation}
L_{\mathrm{Rquadratic}} = L_{\mathrm{RInfoNCE}}+ L_{\mathrm{Rgraph}}.
\label{eq:rqua}
\end{equation}

Note that different loss terms are weighted equally in Eqs.~(\ref{eq:rgraph}) and (\ref{eq:rqua}), and our method performs well without laborious searching on the balance hyper-parameters.

\section{Experiments}

In this section, we carry out extensive experiments on three widely-used GM datasets with the comparisons of 12 state-of-the-art deep graph matching approaches.

\subsection{Experimental Settings}
 
\paragraph{Datasets.} We conduct experiments on Pascal VOC with Berkeley annotation~\cite{pascalvoc}, SPair-71K~\cite{spair71k}, and Willow Object Class~\cite{willow}. For extensive evaluation, we report the average and per-category performance. The objective of all experiments is to maximize the average matching accuracy.

\paragraph{Implementation details.}
We implement our method in PyTorch 1.10.0 and conduct all evaluations on the Ubuntu-20.04 OS with an NVIDIA 3090 GPU. 
For all datasets, we use the exact same set of hyper-parameters. 
The encoder network consists of an ImageNet-pretrained VGG16~\cite{vgg} image encoder, a graph neural network SplineCNN~\cite{splinecnn}, and a two-layer projection head~\cite{simclr}. 
All details of the network have been presented in Appendix~\ref{app:b}.
To optimize the networks, we adopt Adam optimizer~\cite{adam} with the default parameters and set the initial learning rate as $3e^{-4}$.
The learning rate for fine-tuning the VGG network is $2e^{-5}$. 
The batch size is set to 8 image pairs and the distillation temperature $\alpha$ is fixed to 0.4. 
To obtain the permutation matrix $\mathbf{Y}$, we perform the Hungarian algorithm on the similarity matrix $\mathbf{S}$ obtained from the base encoder following~\cite{NGM,liu2021joint,PCA,IPCA,ASAR,DLGM}.

\paragraph{Compared methods.} 
We compare our COMMON with the following 12 popular deep graph matching methods: GMN~\cite{GMN}, PCA~\cite{PCA}, NGM~\cite{NGM}, IPCA~\cite{IPCA}, CIE~\cite{CIE}, DGMC~\cite{DGMC}, LCS~\cite{LCS}, BBGM~\cite{BBGM}, QC-DGM~\cite{QCDGM}, NGM-v2~\cite{NGM}, SCGM~\cite{selfgm}, and ASAR~\cite{ASAR} on the popular open-source graph matching toolbox ThinkMatch\footnote{\href{https://github.com/Thinklab-SJTU/ThinkMatch}{https://github.com/Thinklab-SJTU/ThinkMatch}}, for the purpose of better reproducibility and more fair comparison. Note that our code has also been included in ThinkMatch.

\begin{table*}
\tablestyle{2.6pt}{1.01}
{
\begin{tabular}{l|cccccccccccccccccc|c}
\toprule

Method  & {  Aero} & {  Bike} & {  Bird} & {  Boat} & {  Bottle} & {  Bus} & {  Car} & {  Cat} & {  Chair} & {  Cow} & {  Dog} & {  Horse} & {  Mbike} & {  Person} & {  Plant} & {  Sheep} & {  Train} & {  Tv} & Mean \\
\midrule GMN~\cite{GMN} &59.9	&51.0&	74.3&	46.7&	63.3&	75.5	&69.5	&64.6	&57.5	&73.0	&58.7	&59.1&	63.2	&51.2	&86.9	&57.9	&70.0	&92.4&	65.3  \\
PCA~\cite{PCA}& 64.7&	45.7&	78.1	&51.3	&63.8	&72.7&	61.2&	62.8	&62.6&	68.2	&59.1	&61.2	&64.9&	57.7&	87.4	&60.4	&72.5	&92.8&	66.0 \\
NGM~\cite{NGM} & 66.4&	52.6&	77.0&	49.6&	67.7	&78.8&	67.6&	68.3	&59.2&	73.6&	63.9&	60.7	&70.7&	60.9	&87.5&	63.9	&79.8	&91.5&	68.9 \\
IPCA~\cite{IPCA} &69.0&	52.9	&80.4&	54.3	&66.5&	80.0&	68.5&	71.4&	61.4	&74.8	&66.3&	65.1	&69.6&	63.9&	91.1	&65.4	&82.9	&97.5	&71.2  \\
CIE~\cite{CIE}& 71.5&	57.1&	81.7	&56.7	&67.9&	82.5	&73.4&	74.5	&62.6	&78.0&68.7&	66.3	&73.7&	66.0&	92.5&	67.2&	82.3&	97.5	&73.3 \\
NGM-v2~\cite{NGM}&68.8&
63.3&	86.8	&70.1	&69.7&	94.7	&87.4	&77.4&	72.1	&80.7	&74.3	&72.5&	79.5	&73.4	&98.9	&81.2	&94.3	&98.7	&80.2 \\
BBGM~\cite{BBGM} &\underline{75.3} &	\underline{65.0} &	87.6 &	78.0 &	69.8 &	94.0 	&87.8 &	\underline{78.3}& 	\underline{72.8} 	&82.7 &	76.6 	&\underline{76.3} &	80.1 &	75.0& 	98.7 	&\textbf{85.2} &	96.3 &	98.0 &	82.1 \\
ASAR~\cite{ASAR} 
&	72.4	&61.8&	\underline{91.8}	&\underline{79.1}	&\textbf{71.2}&	\underline{97.4}	&\underline{90.4}&	\underline{78.3}	&\textbf{74.2}&	\underline{83.1}&	\underline{77.3}	&\textbf{77.0}	&\underline{83.1}	&\underline{76.4}	&\underline{99.5}	&\textbf{85.2}	&\underline{97.8}	&\underline{99.5}	&\underline{83.1}
\\
\midrule
COMMON &
\textbf{77.3}&                                                                         
\textbf{68.2}&
\textbf{92.0}&  
\textbf{79.5}& 
\underline{70.4}&  
\textbf{97.5}&    
\textbf{91.6}&    
\textbf{82.5}&     
72.2&   
\textbf{88.0}&     
\textbf{80.0}&     
74.1&  
\textbf{83.4}&                                                                              
\textbf{82.8}&  
\textbf{99.9}&                                                                         
84.4 &
\textbf{98.2}&
\textbf{99.8}                                                                          
&\textbf{84.5}

\\ \bottomrule
\end{tabular}
}
\vspace{0pt} 
\caption{
 Keypoint matching accuracy (\%) on SPair-71k for all classes. }
\label{tab:spair71k}
\end{table*}

\begin{figure*}
  \centering
  \subfigure[Performance with varying noise ratios.]
  {\includegraphics[width=0.32\linewidth]{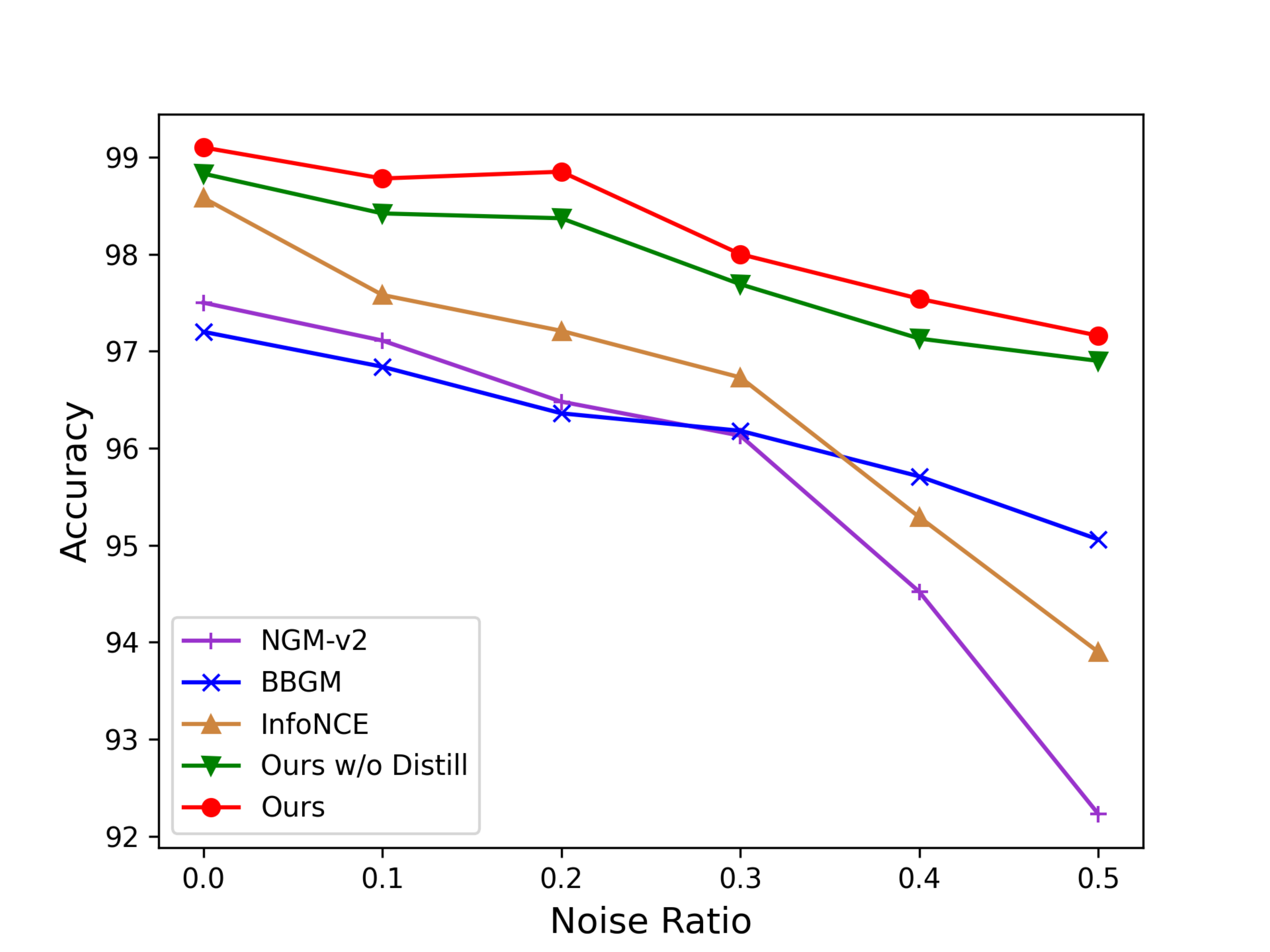}
    \label{fig:nc-a}
    }
  \subfigure[Initial distribution of similarity.]
  {\includegraphics[width=0.32\linewidth]{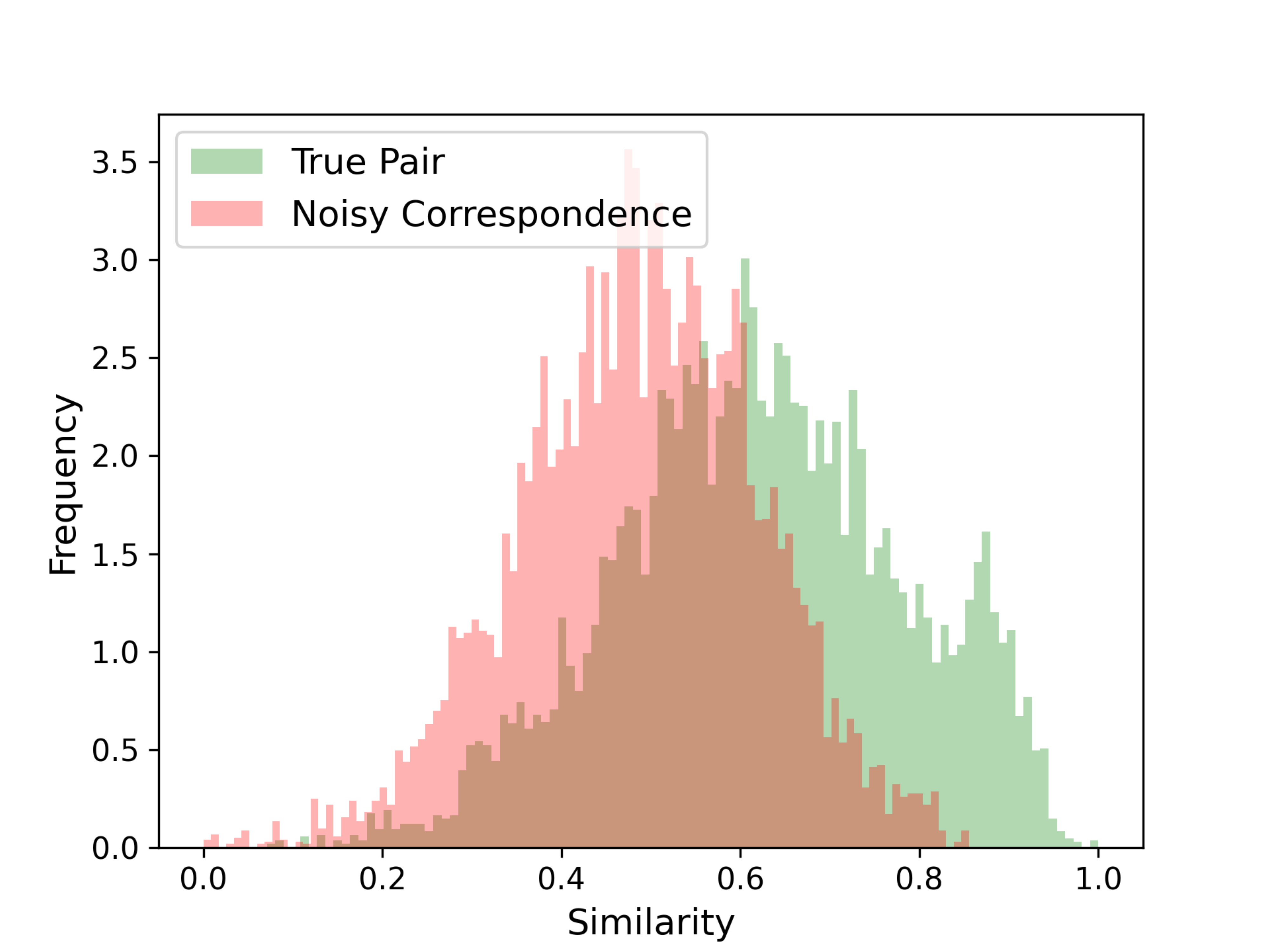}
    \label{fig:nc-b}
    }
  \subfigure[Similarity distribution after training.]
   {\includegraphics[width=0.32\linewidth]{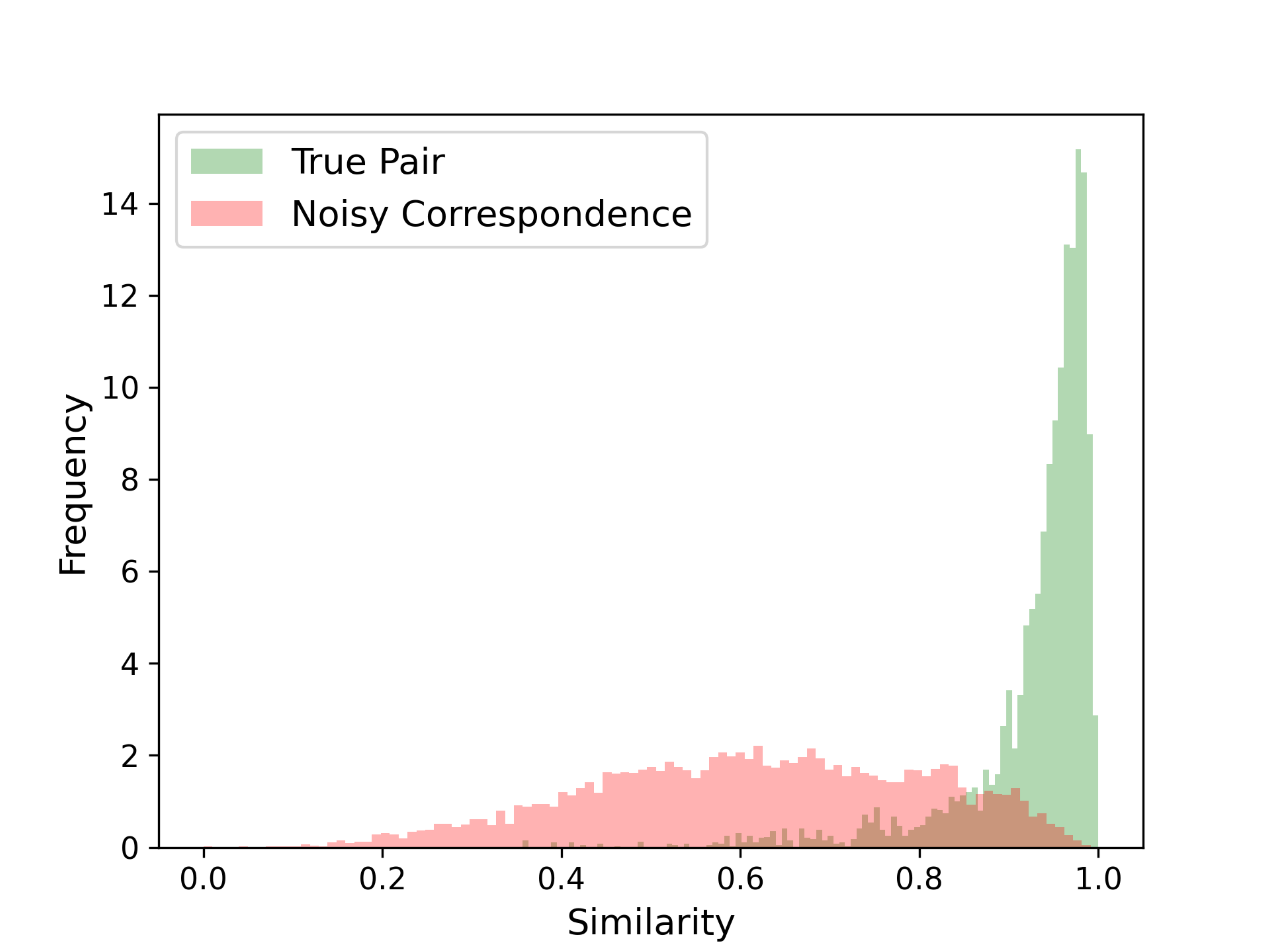}
    \label{fig:nc-c}}
    \vspace{3pt}
  \caption{Effectiveness on noisy correspondence.}
  \label{fig:nc_analy}
\end{figure*}

\subsection{Results on Graph Matching}

\paragraph{Pascal VOC}~\cite{pascalvoc} consists of 7,020 training images and 1,682 testing images with 20 classes in total. The number of nodes per graph ranges from 6 to 23. 
Following~\cite{BBGM}, we test our method by filtering out the non-matched points before matching.
As shown in Table~\ref{tab:pascal}, our method outperforms the counterparts with \plus1.6\% in terms of accuracy. It is worth pointing out that our method achieves remarkable performance improvements on the classes with prominent noisy correspondence, \textit{e.g.,} table (\plus6.8\%) and sofa (\plus3.8\%).

\paragraph{SPair-71k}~\cite{spair71k}  consists of 70,958 image pairs collected from Pascal VOC 2012 and Pascal 3D\plus. 
 Following the data preparation in~\cite{NGM,PCA,IPCA}, each object is cropped to its bounding box and scaled to 256$\times$256. As shown in Table~\ref{tab:spair71k}, our method consistently improves the matching performance (\plus1.4\%). 

\paragraph{Willow Object}~\cite{willow} consists of 256 images in 5 categories, where each target object is annotated with 10 distinctive landmarks. Following the protocol in~\cite{PCA,IPCA,NGM}, we train our methods on the first 20 images and report testing results on the rest. 
As shown in Table \ref{tab:willow_detail}, our method significantly outperforms baselines (\plus 1.4\%).

\begin{figure*}
  \centering
\includegraphics[width=1\linewidth]{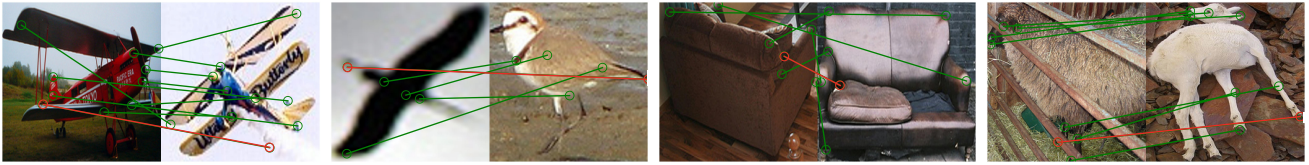}
   \caption{\textbf{Examples of noisy correspondence} in Pascal VOC and SPair-71k. The red line indicates wrongly annotated correspondence. }
  \label{fig:visualization}
\end{figure*}

\subsection{Evaluation on Noisy Correspondence}
In this section, we evaluate the effectiveness of our method against noisy correspondence with synthetic noise experiments, varying viewpoint experiments, ablation studies, and parameter analysis.

\subsubsection{Synthetic Noise Experiment}
To explicitly demonstrate the effectiveness of our method against noisy correspondence, we evaluate models on Willow Object with synthetic noisy correspondence. 
Concretely, we randomly select some keypoints in the training set as the noise points by adding displacement to the location coordinates. The displacement $(s,\theta)$ is generated from uniform distribution: $s \sim \mathcal{U}(0.1,0.2),~ \theta \sim  \mathcal{U}(0,360)$ where $s$ is the displacement value and $\theta$ is the angle. The displacement value is further scaled with respect to the size of the bounding box. The noisy rate $\eta$ is defined as the percentage of noise points.

\paragraph{Varying noise ratios.} As shown in Fig.~\ref{fig:nc-a}, we conduct experiments by varying the noise rate $\eta$ from 0 to 0.5 with an interval of 0.1. From the results, one could observe that: i) our method significantly outperforms baselines in all noise rate settings; ii) momentum distillation consistently improves the robustness against noisy correspondence.

\paragraph{Distribution of similarity scores.}
We further explore the similarity scores of the correspondence learned by our method under noise rate $\eta=0.3$. 
In Figs.~\ref{fig:nc-b} and \ref{fig:nc-c}, we show the histograms of similarity scores for true pairs (in which keypoints are without synthetic noise) and noisy correspondence (at least one of the keypoints is with synthetic noise) separately.
The initial similarity of these two kinds of pairs is confused and hardly distinguishable. 
After training, the similarity of noisy correspondence is 0.6 on average, while that of true pairs is 0.95 on average and ranges into $[0.7,~1.0]$. In other words, our method prevents noisy correspondence from dominating the network optimization, thus eliminating the negative impact of noisy correspondence.

\subsubsection{Evaluation with Different Viewpoint Difficulty} 

Spair-71k contains images with diverse variations in viewpoint and divides image pairs into easy, medium, and hard ones. In practice, we find that noisy correspondence such as occlusion is more likely to occur in hard image pairs. In other words, with the increment of viewpoint difficulty, noisy correspondence appears more frequently. As shown in Table~\ref{tab:viewpoint}, our method consistently improves the matching result, particularly for the image pairs with high viewpoint difficulty (\plus2.8\% for hard pairs). This experiment highlights the ability of our method against noisy correspondence.

\begin{table}[t]
\tablestyle{5pt}{1.02}
{
\begin{tabular}{l|ccccc|c}
\toprule Method & Car & Duck & Face & Mbike &Wbottle& Mean \\
\midrule 
GMN~\cite{GMN}     & 67.9 & 76.7 & 99.8  & 69.2      & 83.1       & 79.3 \\
NGM~\cite{NGM}     & 84.2 & 77.6 & 99.4  & 76.8      & 88.3       & 85.3 \\
PCA~\cite{PCA}  & 87.6 & 83.6 &  \textbf{100} & 77.6      & 88.4       & 87.4 \\
CIE~\cite{CIE}   & 85.8 & 82.1 & 99.9  & 88.4      & 88.7       & 89.0 \\
IPCA~\cite{IPCA} & 90.4 & 88.6 &  \textbf{100} & 83.0      & 88.3       & 90.1 \\
SCGM~\cite{selfgm}&91.3 &73.0 &100 &95.6 &96.6 &91.3 \\
ASAR~\cite{ASAR} & 92.5&84.0&\textbf{100}&95.4&{99.0}&94.2 \\
LCS~\cite{LCS} &91.2 &86.2 &\textbf{100} &99.4 &97.9 &94.9 \\ 
DGMC~\cite{DGMC}  & \textbf{98.3} & 90.2 & \textbf{100}& 98.5 & 98.1 & 97.0 \\
BBGM~\cite{BBGM}    & 96.8 & 89.9 &  \textbf{100} & \underline{99.8}      & \underline{99.4}       & 97.2 \\
NGM-v2~\cite{NGM}  & 97.4 &\underline{93.4} &  \textbf{100} & 98.6      & 98.3       & 97.5\\
QC-DGM~\cite{QCDGM} & \underline{98.0} & 92.8 & \textbf{100} & 98.8 & {99.0} & \underline{97.7} \\
\midrule 
COMMON &
97.6& \textbf{98.2}&\textbf{100} &\textbf{100} &\textbf{99.6} & \textbf{99.1}
\\
\bottomrule

\end{tabular}
}
\vspace{4pt} 
\caption{{Keypoint matching accuracy} (\%) across all objects on Willow Object. }
\label{tab:willow_detail}
\end{table}

\begin{table}[t]
\tablestyle{4.5pt}{1.01}
{
\centering
\begin{tabular}{lcccc}
\toprule \multirow{2}{*}{ Method } & \multicolumn{3}{c}{ Viewpoint difficulty } & \multirow{2}{*}{ All } \\
& Easy & Medium & Hard & \\
\midrule

BBGM~\cite{BBGM} & 84.7  & 78.9&73.6& 82.1 \\
ASAR~\cite{ASAR} 
&86.5&79.1&72.5&83.1\\
\midrule
COMMON & \textbf{86.6}~(\plus0.1) & \textbf{ 81.4}~(\plus2.3) & \textbf{76.4}~(\plus2.8) & \textbf{84.5}~(\plus 1.4)\\

\bottomrule
\end{tabular}
}
\vspace{0pt} 
\caption{Keypoint matching accuracy (\%) on SPair-71k grouped by levels of difficulty in the viewpoint of the matching pair.  }
\label{tab:viewpoint}
\end{table}

\subsubsection{Ablation Studies and Parameter Analysis}

To evaluate our framework, we conduct comprehensive ablation studies by separately investigating each component. As shown in Table~\ref{tab:ablation}, all modules are inseparable and bring substantial performance gains. Notably, the momentum distillation strategy remarkably improves the performance by alleviating the negative impact of bi-level noisy correspondence. Besides,  the graph consistency terms play indispensable roles in learning quadratic correlations and our loss outperforms the quadratic regularization~\cite{moskalev2022contrasting}.
Following, we present the parameter sensitivity analysis on the distillation temperature $\alpha$ in Table~\ref{tab:parameter}. As shown, the distillation strategy indeed improves the performance and our method is insensitive to the choice of $\alpha$.

\subsection{Visualization on Noisy Correspondence}

In this section, we identify some noisy correspondence pairs predicted by our model with the lowest matching similarity in Fig.~\ref{fig:visualization}. More visualization results have been presented in Appendix~\ref{app:c}.

\section{Conclusion}

\begin{table}[t]
\tablestyle{4pt}{1.01}
{
\centering
\begin{tabular}{lcc}
\toprule {Method} &  Pascal VOC &SPair-71k \\ 
\midrule
{\textbf{COMMON} (FULL)}&\baseline{\textbf{82.67}} &\baseline{\textbf{84.54}}   \\
- w/o graph consistency &81.95&83.83\\

- w/o distillation & 81.77 & {83.94}\\

\midrule
- InfoNCE and quadratic regularization~\cite{moskalev2022contrasting} & 81.67 &83.66 \\
- InfoNCE  &81.33& 83.38 \\

\bottomrule
\end{tabular}
}
\caption{\textbf{Ablation study} of COMMON in terms of accuracy. Default settings are marked in \colorbox{baselinecolor}{gray}.
}
\label{tab:ablation}
\end{table}

\begin{table}[t]
\tablestyle{1.8pt}{1.01}
{
\centering
\begin{tabular}{c|cccccccccc}
\toprule 
{$\alpha$} &  0 & 0.1& 0.2&0.3&\baseline{0.4}&0.5&0.6&0.7&0.8&0.9 \\ 
\midrule
ACC&{81.90}	&82.34&	\textbf{82.69}&	82.62	&\baseline{82.67}	&82.50	&82.59	&82.62&	82.66&	82.41 
\\
\bottomrule
\end{tabular}
}
\caption{\textbf{Parameter analysis} of COMMON with the increase of the distillation parameter $\alpha$ on Pascal VOC. Default settings are marked in \colorbox{baselinecolor}{gray}.
}
\label{tab:parameter}
\end{table}

This paper reveals a new problem for graph matching, \textit{i.e.}, bi-level noisy correspondence, which refers to wrongly annotated node-to-node correspondence and accompanied edge-to-edge correspondence. To tackle this challenge, the proposed method introduces the momentum distillation strategy to rebalance the quadratic contrastive loss and mitigate the influence of BNC. This work potentially provides a novel insight into the community and may inspire some further exploration of the noisy correspondence problem. In terms of application scenario, noisy correspondence might appear in not only two-graph matching but also multi-graph and hyper-graph matching. In terms of methodology, it may be feasible to design NC-robust GM networks by controlling the information propagation in graph neural networks. 

\section*{Acknowledgments}

This work was supported by the National Key R\&D Program of China under Grant 2020YFB1406702, in part by NSFC under Grant U21B2040 and 62176171, and in part by the 111 Project under grant B21044.

\appendix

\section*{Appendix}
\label{sec:intro}

In this appendix, we first present the proof of Theory~\ref{the:1}. After that, we present experiment details about the network architectures and more experiment results to further investigate the effectiveness of our method. Finally, we discuss the broader impact of our work.

\setcounter{Theo}{0}
\section{Proof of Theorem 1}
\label{app:a}

This theorem is based on the Proposition 2 of~\cite{moskalev2022contrasting}. We refer the readers to~\cite{moskalev2022contrasting} for more explanations.

\begin{Theo}
The log-sum-exp ~\cite{beck2012smoothing} smoothed structured linear assignment loss $L$ with row-stochastic relaxation is equivalent to the InfoNCE contrastive loss~\cite{moco,mocov2}.
\end{Theo}

\begin{Proof}
We first relax the constraint $\mathbf{Y}\in\Pi$ to $\mathbf{Y}\in\mathcal{R}$ where $\mathcal{R}$ is a set of row-stochastic binary matrix, \textit{i.e.}, $[\mathbf{Y} ]_{ij}\in\{0,1\}$ and $\sum_j\mathbf{Y}_{i j}=1 ~ \forall i$. Based on it, we reformulate the structured linear assignment loss as,
\begin{equation}
  \begin{aligned}
L&=-\operatorname{tr}\left(\mathbf{S} \mathbf{Y}_{g t}^{\top}\right) + \max _{\mathbf{Y} \in \mathcal{R}} \operatorname{tr}(\mathbf{S} \mathbf{Y}^{\top}) \\
&=-\operatorname{tr}\left(\mathbf{S} \mathbf{Y}_{g t}^{\top}\right)+ \max _{y_1 \dots y_n} \sum_i(\sum_j[\mathbf{S}]_{i j}\left[y_i\right]_j) \\
&=-\operatorname{tr}\left(\mathbf{S} \mathbf{Y}_{g t}^{\top}\right)+\sum_i \max _{y_i}(\sum_j[\mathbf{S}]_{i j}\left[y_i\right]_j) \\
&=-\operatorname{tr}\left(\mathbf{S} \mathbf{Y}_{g t}^{\top}\right)+\sum_i \max _j\left[\mathbf{S}_{i j}\right],
\end{aligned}
\end{equation}
where $y_i$ is the $i$-th row of $\mathbf{Y}$. The third identity is based on the independence of the rows $y_1, \dots,  y_n$ and the last identity follows the fact that $y_i$ is a one-hot vector containing the maximum index. As the structured linear loss is non-smoothness and difficult to optimize, we utilize the common log-sum-exp approximation~\cite{beck2012smoothing} on the max function which leads to,
\begin{equation}
  \begin{aligned}
L =- \sum_{(i, j) \in \mathbf{Y}_{g t}}[\mathbf{S}]_{i j}+\tau \sum_i \log (\sum_j \exp (\frac{1}{\tau}[\mathbf{S}]_{i j})),
\end{aligned}
\label{eq:loglinear}
\end{equation}
where $\tau$ controls the degree of smoothness.
In fact, Eq.~(\ref{eq:loglinear}) is the so-called InfoNCE contrastive loss where the first and second term refer to the alignment and uniformity property~\cite{wang2020understanding,SimCSE}, respectively. 
\hfill $\square$
\end{Proof}

\section{Details of Network Architectures}
\label{app:b}

Given two graphs $\mathcal{G}_A=\left\{\mathbf{U}_A, \mathbf{E}_A\right\}$ and $\mathcal{G}_B=\left\{\mathbf{U}_B, \mathbf{E}_B\right\}$ with $n$ and $m$ keypoints each ($n\leq m$). $\mathbf{U}$ indicates the set of nodes and $\mathbf{E}$ denotes the set of edges. The node and edge features are learned through the base encoder whose structure is nearly the same as that of BBGM~\cite{BBGM} and NGM-v2~\cite{NGM}. Concretely, the base encoder consists of an image encoder, a graph neural network, and a projection head. The proposed momentum encoder is with the same structure as the base encoder.

\paragraph{Image encoder}.
Following~\cite{NGM,QCDGM,BBGM,ASAR,selfgm,PCA,IPCA}, we employ VGG16~\cite{vgg} as the image encoder to extract the node features. 
Specifically, we extract the node features from $\operatorname{relu4\_2}$
 and $\operatorname{relu5\_1}$ of VGG16, and  concatenate them to form the initial node feature matrices ${\bar{\mathbf{U}}}_A \in \mathbb{R}^{n \times d_1}, {\bar{\mathbf{U}}_B} \in \mathbb{R}^{m \times d_1}$ where $d_1=1024$.
 
\paragraph{Graph neural network.}
Following~\cite{NGM,BBGM,selfgm,ASAR},  we initial the edge structure $\mathbf{E}_A \in \mathbb{R}^{n \times n}$ and $\mathbf{E}_B \in \mathbb{R}^{m \times m}$ with Delaunay triangulation and $[\mathbf{E}]_{ij}$ is weighted as the difference between the coordinate positions of keypoint $i$ and $j$.
We pass the initial node features $\bar{\mathbf{U}}$ and the edge structure $\mathbf{E}$ through graph network SplineCNN~\cite{splinecnn}, which is a powerful graph convolution network that encodes geometric features into node features by updating the node representation via a weighted summation of its neighbors. Formally, the update rule at keypoint $i$ is,
\begin{equation}
\operatorname{SplineCNN}  \left([\bar{\mathbf{U}}]_i\right)=
\frac{1}{|\mathcal{N}(i)|}  
\sum_{j \in \mathcal{N}(i)} [\bar{\mathbf{U}}]_{j} \cdot g([\mathbf{E}]_{ij}),
\label{eq:splinecnn}
\end{equation}
where $\mathcal{N}(i)$ indicates the neighbors of node $i$, $g$ is the B-Spline kernel, and $\cdot$ is the dot product.
Separately feeding the graphs $\mathcal{G}_A$ and $\mathcal{G}_B$ into SplineCNN, we obtain the refined node features ${\hat{\mathbf{U}}}_A \in \mathbb{R}^{n \times d_2}, {{\hat{\mathbf{U}}}}_B \in \mathbb{R}^{m \times d_2}$ where $d_2=1024$, respectively. 

\paragraph{Projection Head.} Following classical contrastive learning paradigms~\cite{simclr,mocov2}, we obtain the final node feature $\mathbf{V}_A\in\mathbb{R}^{n\times d_3}$ and $\mathbf{V}_B\in\mathbb{R}^{m\times d_3}$ where $d_3=256$ through two fully-connected layers (FCN). Formally,
\begin{equation}
  \mathbf{V} = \operatorname{norm}\left(
  f_2\left(f_1({\hat{\mathbf{U}}})\right)\right),
\end{equation}
where FCN $f_1$ and $f_2$ are with the batch normalization layer and ReLU activation. $\operatorname{norm}$ operation denotes $\ell_2$ normalization and the dimensionality of $f_1$ and $f_2$ is set to 1024 and 256, respectively.

Finally, we obtain the node similarity matrix $\mathbf{S}$ and the edge adjacency matrices $\mathbf{F}_A,\mathbf{F}_B$ through $\mathbf{S}=\mathbf{V}_A \mathbf{V}_B^{\top}$, $\mathbf{F}_A=\mathbf{V}_A\mathbf{V}_A^{\top}$, and $\mathbf{F}_B=\mathbf{V}_B\mathbf{V}_B^{\top}$.

\begin{figure*}[h]
  \centering
   \includegraphics[width=1\linewidth]{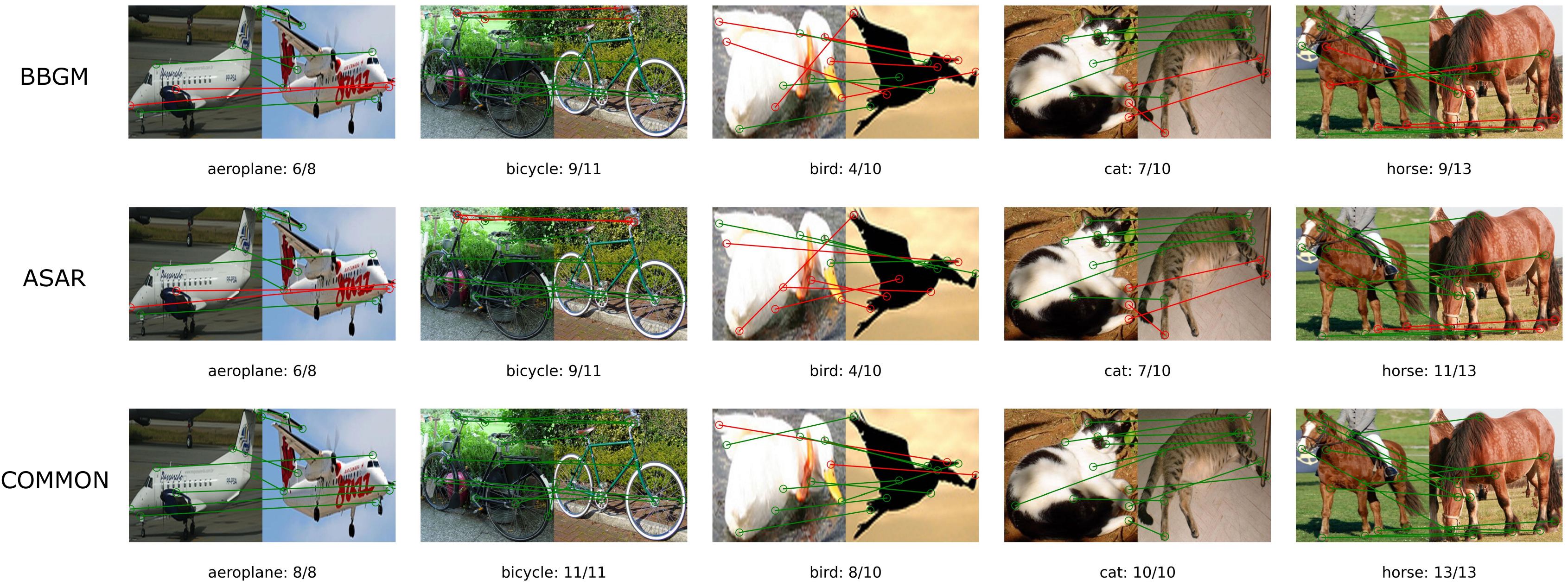}
   \caption{\textbf{Visualization of the matching results} on Pascal VOC. Green and red lines denote correct and false matching results, respectively.
}
   \label{fig:visupascal}
\end{figure*}

\begin{figure*}[!t]
  \centering
   \includegraphics[width=1\linewidth]{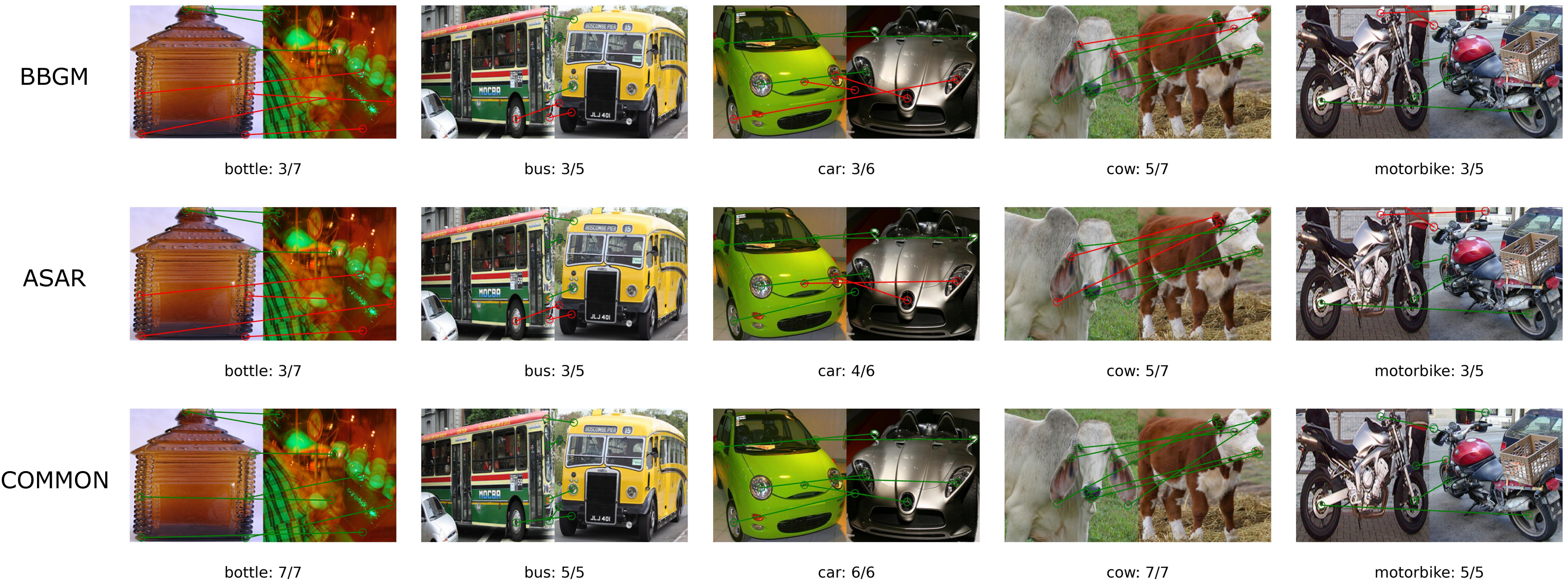}
   \caption{\textbf{Visualization of the matching results} on SPair-71k.
}
   \label{fig:visuspair}
\end{figure*}

\section{Visualization on Graph Matching}
\label{app:c}

We present the visual matching results of our method and the most comparable baselines BBGM~\cite{BBGM} and ASAR~\cite{ASAR} on the Pascal VOC and Spair-71k datasets. For better visualization, we crop the object according to its bounding box.
As shown in Figs.~\ref{fig:visupascal} and~\ref{fig:visuspair}, our method achieves superior matching performance, especially for the image pairs with high viewpoint difficulty and low recognizability.

\section{Broader Impact}
\label{app:d}

This work could be the first work that reveals the importance of the noisy correspondence problem in graph matching. Solving this problem could improve the tolerance for the errors of annotations, which might benefit the practitioners in the industry. 
Although the proposed COMMON achieves remarkable improvement, the complexity of training the model is slightly larger due to the additional momentum network. In practice, we find the time cost is approximately $\x$1.4 times that of training a base encoder only. Fortunately, the inference speed is exactly the same as we only keep the base encoder during testing.

{\small
\bibliographystyle{ieee_fullname}
\bibliography{egbib}
}

\end{document}